%% file: acl2020.tex
\documentclass[11pt,a4paper]{article}
\usepackage[hyperref]{acl2020}
\usepackage{times}
\usepackage{latexsym}
\usepackage{amsfonts}
\usepackage{amsmath}
\usepackage{graphicx}
\usepackage{xspace}
\usepackage{multirow}
\usepackage{url}

\usepackage{microtype}
\graphicspath{ {./images/} }
\aclfinalcopy %

\newcommand{\norm}[1]{\left\lVert #1 \right\rVert}

\newcommand{\atis}{\textsc{ATIS}\xspace}

\newcommand{\snips}{\textsc{Snips}\xspace}
\newcommand{\maml}{\textsc{MAML}\xspace}
\newcommand{\fbtop}{\textsc{TOP}\xspace}
\newcommand{\lstm}{\textsc{LSTM}\xspace}
\newcommand{\elmo}{\textsc{ELMo}\xspace}
\newcommand{\bert}{\textsc{BERT}\xspace}
\newcommand{\glove}{\textsc{GloVe}\xspace}
\newcommand{\ft}{\textsc{Fine-tune}\xspace}
\newcommand{\fomaml}{fo\textsc{MAML}\xspace}
\newcommand{\Lagr}{\mathcal{L}}
\newcommand{\slot}{slot filling\xspace}
\newcommand{\slt}{SF\xspace}

\title{Learning to Classify Intents and Slot Labels Given a Handful of Examples}

\author{Jason Krone \\
  Amazon AI \\ Palo Alto, CA \\
  {\tt kronej@amazon.com} \\\And
  Yi Zhang \\
  Amazon AI \\ Seattle, WA \\
  {\tt yizhngn@amazon.com} \\\And
  Mona Diab\Thanks{Work performed while at Amazon AI} \\
  The George Washington University \\ 
  Washington, DC \\
  {\tt mtdiab@gwu.edu} \\
}

\date{}

\begin{document}
\maketitle

\input{sections/abstract}

\input{sections/introduction}

\input{sections/related_work}

\input{sections/task}

\input{sections/approach}

\input{sections/benchmark}

\input{sections/experiments}

\input{sections/tables}

\input{sections/results}

\input{sections/conclusion}

\bibliographystyle{acl_natbib}
\bibliography{anthology,acl2020}

\end{document}

%% file: sections/abstract.tex
\begin{abstract}
Intent classification (IC) and \slot (\slt) are core components in most goal-oriented dialogue systems.
Current IC/\slt models perform poorly when the number of training examples per class is small. 
We propose a new few-shot learning task, \emph{few-shot IC/\slt}, to study and improve the performance of IC and \slt models on classes not seen at training time in ultra low resource scenarios. 
We establish a \emph{few-shot IC/\slt} benchmark by defining few-shot splits for three public IC/\slt datasets, \atis, \fbtop, and \snips. 
We show that 
two popular few-shot learning algorithms, model agnostic meta learning (\maml) and prototypical networks, outperform a fine-tuning baseline on this benchmark. Prototypical networks achieves significant gains in IC performance on the \atis and \fbtop datasets, while both prototypical networks and \maml outperform the baseline with respect to \slt on all three datasets. 
In addition, we demonstrate 
that joint training as well as the use of pre-trained language models, ELMo and BERT in our case, are complementary to these few-shot learning methods and yield further gains. 
\end{abstract}

%% file: sections/introduction.tex
\section{Introduction}
In the context of goal-oriented dialogue systems, 
intent classification (IC) is the process of classifying a user's utterance into an intent, such as \emph{BookFlight} or \emph{AddToPlaylist}, referring to the user's goal. While slot filling (\slt) is the process of identifying and classifying certain tokens in the utterance into their corresponding labels, in a manner akin to named entity recognition (NER). However, in contrast to NER, typical slots are particular to the domain of the dialogue, such as music or travel. As a reference point, we list intent and slot label annotations for an example utterance from the \snips dataset with the \emph{AddToPlaylist} IC in Figure \ref{snipsexample}.

\begin{figure}
\begin{center}
\small
\begin{tabular}{c c}
\hline
Token & Slot Label \\
\hline
Please & O \\
add & O  \\
some & O  \\
Pete & AddToPlaylist:artist  \\
Townshend & AddToPlaylist:artist  \\
to & O  \\
my & AddToPlaylist:playlist\_owner  \\
playlist & O  \\
Fiesta & AddToPlaylist:playlist  \\
Hits  & AddToPlaylist:playlist  \\
con & AddToPlaylist:playlist  \\
Lali & AddToPlaylist:playlist  \\
\hline
\end{tabular}
\caption{Tokens and corresponding slot labels for an utterance from the AddToPlaylist intent class in the \snips dataset prefixed by intent class name.}
\label{snipsexample}
\end{center}
\end{figure}

As of late, most state-of-the-art IC/\slt models are based on feed-forward, convolutional, or recurrent neural networks \cite{hakkani2016multi, goo2018slot, gupta2019simple}.
These neural models offer substantial gains in performance, but they often require a large number of labeled examples (on the order of hundreds) per intent class and slot-label to achieve these gains.
The relative scarcity of large-scale datasets annotated with intents and slots prohibits the use of neural IC/\slt models in many promising domains, such as medical consultation, where it is difficult to obtain large quantities of annotated dialogues. 

Accordingly, we propose the task of \emph{few-shot IC/\slt}, catering to domain adaption in low resource scenarios, where there are only a handful of annotated examples available per intent and slot in the target domain.
To the best of our knowledge, this work is the first to apply the few-shot learning framework to a joint sentence classification and sequence labeling task.
In the NLP literature, few-shot learning often refers to a low resource, cross lingual setting where there is limited data available in the target language. 
We emphasize that our definition of \emph{few-shot IC/\slt} is distinct in that we limit the amount of data available per target class rather than target language.

\emph{Few-shot IC/\slt} builds on a large body of existing few-shot classification work. %
Drawing inspiration from computer vision, 
we experiment with two prominent few shot image classification approaches, \emph{prototypical networks} and \emph{model agnostic meta learning} (\maml). %
Both these methods seek to decrease over-fitting and improve generalization on small datasets, albeit via different mechanisms. \emph{Prototypical networks} learns class specific representations, called \emph{prototypes}, and performs inference by assigning the class label associated with the prototype closest to an input embedding. Whereas \emph{\maml} modifies the learning objective to optimize for pre-training representations that transfer well when fine-tuned on a small number of labeled examples.

For benchmarking purposes, we establish few-shot splits for three publicly available IC/\slt datasets: \atis \cite{hemphill1990atis}, \snips \cite{coucke2018snips}, and \fbtop \cite{gupta-etal-2018-semantic-parsing}.
Empirically, 
prototypical networks yields substantial improvements on this benchmark over the popular \textit{``fine-tuning''} approach \cite{goyal2018fast, schuster2018cross}, where representations are pre-trained on a large, 
\textit{``source''} dataset and then fine-tuned on a smaller, \textit{``target''} dataset.
Despite performing worse on intent classification, \maml also achieves gains over \textit{``fine-tuning''} on the \slot task.
Orthogonally, we experiment with the use of two pre-trained language models, \bert and \elmo, as well as joint training on multiple datasets. 
These experiments show that the use of pre-trained, contextual representations is complementary to both methods. While prototypical networks is uniquely able to leverage joint training to consistently boost \slot performance.

In summary, our primary contributions are four-fold:
\begin{enumerate}
\item 
Formulating IC/\slt as a few-shot learning task;
\item
Establishing few-shot splits for the \atis, \snips, and \fbtop datasets;
\item
Showing that \maml and prototypical networks can outperform the popular ``fine-tuning" domain adaptation framework;
\item
Evaluating the complementary of contextual embeddings and joint training with \maml and prototypical networks.

\end{enumerate}

%% file: sections/related_work.tex
\section{Related Work}

\subsection{Few-shot Learning}
Early adoption of few-shot learning in the field of computer vision has yielded promising results. Neural approaches to few-shot learning in computer vision fall mainly into three categories: \emph{optimization-}, \emph{metric-}, or \emph{memory-based}.
\emph{Optimization-based} methods typically learn an initialization or fine-tuning procedure for a neural network.
For instance, \maml \cite{finn2017model} %
directly optimizes for representations that generalize well to unseen classes given a few labeled examples.
Using an \lstm based meta-learner, \citet{ravi2016optimization} learn both the initialization and the fine-tuning procedure.
In contrast, \emph{metric-based} approaches learn an embedding space or distance metric under
which examples belonging to the same class have high similarity.
Prototypical networks \cite{snell2017prototypical}, siamese neural networks \cite{koch2015siamese}, and matching networks \cite{vinyals2016matching} all belong to this category. 
Alternatively, memory based approaches apply memory modules or recurrent networks with memory, such as a LSTM, to few-shot learning.
These approaches include differentiable extensions to k-nearest-neighbors \cite{kaiser2017learning} and applications of the Neural Turing Machines \cite{graves2014neural,santoro2016meta}.

\subsection{Few-shot Learning for Text Classification}
To date, applications of few-shot learning to natural language processing focus primarily on text classification tasks.
\citet{yu2018diverse} identify \textit{``clusters''} of source classification tasks that transfer well to a given target task, and meta learn a linear combination of similarity metrics across \textit{``clusters''}.
The source tasks with the highest likelihood of transfer are used to pre-train a convolutional network that is subsequently fine-tuned on the target task.
\citet{2018fewrel} propose \emph{FewRel}, a few-shot relation classification dataset, and use this data to benchmark the performance of few-shot models, such as \emph{prototypical networks} and \emph{SNAIL} \cite{mishra2017simple}.
\emph{ATAML} \cite{jiang2018attentive}, one of the few optimization based approaches to few-shot sentence classification, extends \emph{\maml} to learn task-specific as well as task agnostic representations using feed-forward attention mechanisms. %
\cite{dou2019investigating} show that further pre-training of contextual representations using \emph{optimization-based} methods benefits downstream performance. 

\subsection{Few-shot Learning for Sequence Labeling}
In one of the first works on few-shot sequence labeling, \citet{fritzler2019few} apply prototypical networks to few-shot named entity recognition by
training a separate prototypical network for each named entity type.
This design choice makes their extension of prototypical networks more restrictive than ours, which trains a single model to classify all sequence tags.
\cite{hou2019few} apply a CRF based approach that learns emission scores using pre-trained, contextualized embeddings to few-shot \slt (on \snips) and few-shot NER.

%% file: sections/task.tex
\section{Task Formulation}

\subsection{Few-shot Classification}\label{sec:fewshotdef}
The goal of \emph{few-shot classification} is to adapt a classifier $f_\phi$ to a set of new classes $L$ not seen at training time, given a \emph{few} labeled examples per class $l\in L$. 
In this setting, train and test splits are defined by disjoint class label sets $L_{train}$ and $L_{test}$, respectively. The classes in $L_{train}$ are made available for pre-training and those in $L_{test}$ are held out for low resource adaptation at test time.
Few-shot evaluation is done episodically, i.e. over a number of mini adaptation datasets, called episodes.
Each episode consists of a \emph{support set $S$} and a \emph{query set $Q$}.
The \emph{support set} contains $k_l$ labeled examples $S_l = \{(x_l^i, y_l)|i{\in}(1{\ldots}k_l)\}$ per held out class $l \in L$; we define $S = \bigcup_{l \in L} S_l$.
Similarly, the \emph{query set} contains $k_q$ labeled instances $Q_l = \{(x_l^j,y_l)|j{\in}(1{\ldots}k_q)\}$ for each class $l \in L$ s.t. $Q_l \cap S_l = \{ \}$; we define $Q = \bigcup_{l \in L} Q_l$.
The support set provides a few labeled examples of new classes not seen at training time that $f_\phi$ must adapt to i.e. learn to classify, whereas the query set is used for evaluation.
Few-shot classification requires episodic evaluation;
however, most few-shot learning methods train as well as evaluate on episodes. Consistent with prior work, we train both \maml and prototypical networks methods on episodes, as opposed to mini-batches.

\subsection{Few-shot IC/\slt}
\emph{Few-shot IC/SF} extends the prior definition of \emph{few-shot classification} to include both IC and \slt tasks. 
As \citet{geng2019few} showed, it is straightforward to formulate IC as a  \emph{few-shot classification} task.
Simply let the class labels $y_l$ in section \ref{sec:fewshotdef} correspond to IC labels and partition the set of ICs into the train and test splits, $L_{train}$ and $L_{test}$.
Building on this few-shot IC formulation, we re-define the \emph{support} and \emph{query} sets to include the slots $t_l$, in addition the intent $y_l$, assigned to each example $x_l$.
Thus, the set of \emph{support} and \emph{query} instances for class $l \in L$ become $S_l = \{(x_l^i, t_l^i, y_l)|i{\in}(1{\ldots}k_l)\}$ and $Q_l = \{(x_l^j, t_l^j, y_l)|j{\in}(1{\ldots}k_q)\}$, respectively.
To construct an episode, we sample a total of $k_l + k_q$ labeled examples per IC $l \in L$ to form the support and query sets. 
Since many slot-label sequences may belong to the same IC, 
it is possible to sample an episode such that a slot-label in the query set does not appear in the support set or vice versa.
Therefore, to ensure fair evaluation, we map any slot-label in the query set that does not occur in the support set or vice versa to ``Other", which is ignored by our \slt evaluation metric.

%% file: sections/approach.tex
\begin{figure*}[ht!]
\centering
\includegraphics[scale=0.148]{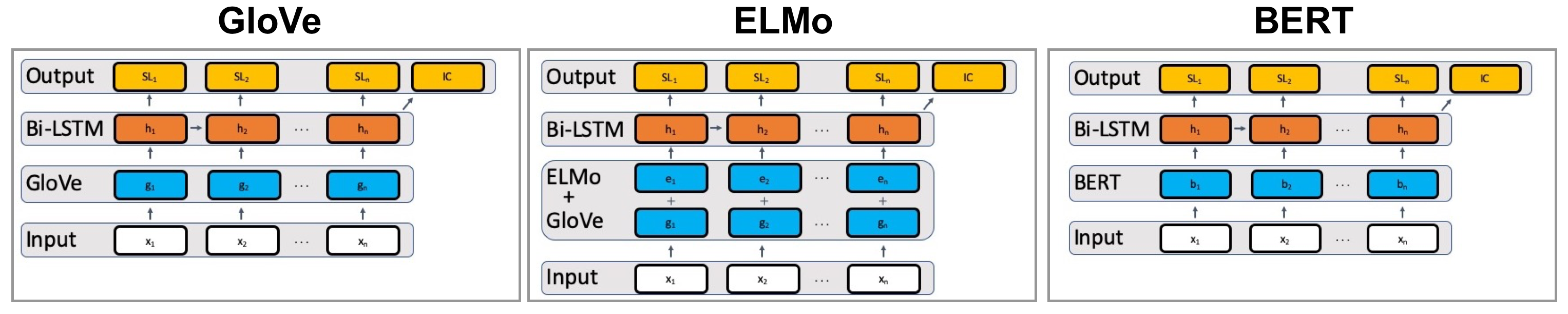}
\caption{Three model architectures, each consisting of an embedding layer, comprised of either GloVe word embeddings (GloVe), GloVe word embeddings concatenated with ELMo embeddings (ELMo), or BERT embeddings (BERT), that feed into a bi-directional LSTM, which is followed by fully connected intent and slot output layers.}
\label{fig:architecture}
\end{figure*}

\section{Approach}

\subsection{Prototypical Networks for Joint Intent Classification and Slot Filling}
The original formulation of prototypical networks \cite{snell2017prototypical} is not directly applicable to sequence labeling.
Accordingly, we extend prototypical networks to perform joint sentence classification and sequence labeling.
Our extension computes ``prototypes" $c_l$ and $c_a$ for each intent class $l$ and slot-label $a$,  respectively.
Each prototype $c \in \mathbb{R^D}$ is the mean vector of the embeddings belonging to a given intent class or slot-label class.
These embeddings are output by a sequence encoder $f_{\phi}(x): \rightarrow \mathbb{R}^D$, which takes a variable length utterance of $m$ tokens $x^i=(x^i_1, x^i_2, \dots, x^i_m)$ as input, 
and outputs the final hidden state $h \in \mathbb{R}^D$ of the encoder. 
For ease of notation, 
let $S_l = \{(x_l^i, t_l^i, y_l)\}$ be the support set instances with intent class $y_l$.
And let $S_a = \{(x^i_{[1:j]}, t^i_{[1:j]}, y^i) | t^i_j = a\}$ be the support set sub-sequences with slot-label $a$ for the token $x^i_j$ in $x^i$. 
Using this notation, we calculate slot-label and intent class prototypes as follows:

\begin{align}
c_l &= \frac{1}{|S_l|} \sum_{(x^i, t^i, y_l) \in S_l} f_{\phi}(x^i) \\
c_a &= \frac{1}{|S_a|} \sum_{(x^i_{[1:j]}, \: t^i_{[1:j]}, \: y^i)} f_{\phi}(x^i_{[1:j]})
\end{align}

Given an example $(x^*, t^*, y^*) \in Q$, we compute the conditional probability $p(y = l \mid x^*, S)$ 
that the utterance $x^*$ has intent class 
$l$
as the normalized Euclidean distance between $f_{\phi}(x^*)$ and the prototype $c_l$,

$$p(y=l \mid x^*, S) = \frac{ \exp( - \norm{f_{\phi}(x^*) - c_{l}}_2^2)}{\sum_{l'} \exp(- \norm{f_{\phi}(x^*) - c_{l'}}_2^2)}$$

Similarly, we compute the conditional probability $p( t^*_j = a \mid x^*, S)$ 
that the j-th token $x^*_j$ in the utterance $x^*$
has slot-label $t^*_j = a$
as the normalized Euclidean distance between $f_{\phi}(x^*_{[1:j]})$ and the prototype $c_a$,

$$p(t^*_j=a \mid x^*, S) = \frac{\exp( - \norm{f_{\phi}(x^*_{[1:j]}) - c_{a}}_2^2)}{\sum_{a'}  \exp(- \norm{f_{\phi}(x^*_{[1:j]}) - c_{a'}}_2^2)}$$

We define the joint IC and \slt prototypical loss function $\Lagr_{proto}$ as the sum of the IC and \slt negative log-likelihoods averaged over the query set instances given the support set: 
\begin{align}
&\Lagr_{proto} = \frac{1}{|Q|} \sum_{(x^*, t^*, y^*) \in Q} \Lagr_{protoIC} + \Lagr_{proto\slt} \nonumber \\
&\Lagr_{protoIC} =  -\log p(y=y^* \mid x^*, S)  \nonumber \\
&\Lagr_{proto\slt} = -\sum_{t^*_j \in t^*} \log p(t^*_j=a \mid x^*, S) \nonumber 
\end{align}

\subsection{Model Agnostic Meta Learning (\maml)}
\maml optimizes the parameters $\phi$ of the encoder $f_\phi$ such that when $\phi$ is fine-tuned on the support set $S$ for $d$ steps,  $\phi' \leftarrow \mbox{Finetune}(\phi, d \:| S)$, the fine-tuned model $f_{\phi'}$ generalizes well to new class instances in the query set $Q$. 
This is achieved by updating $\phi$ to minimize the loss of the fine-tuned model $\Lagr(f_{\phi'}, Q)$ on the query set $Q$.
The update to $\phi$ takes the form $\phi \leftarrow \phi - \nabla_\phi \Lagr(f_{\phi'}, Q)$, where
 $\Lagr$ is the sum of IC and \slt softmax cross entropy loss functions.
Concretely, given a support and query set $(S, Q)$, \maml performs the following two step optimization procedure:
$$1. \:\: \phi' \leftarrow \mbox{Finetune}(\phi, d \:| S) $$
$$2. \:\: \phi \leftarrow \phi - \nabla_\phi \Lagr(f_{\phi'}, Q) $$
Although, the initial formulation of \maml, which we outline here, utilizes stochastic gradient descent (SGD) to update the initial parameters $\phi$, in practice, an alternate gradient based update rule can be used in place of SGD. Empirically, we find it beneficial to use Adam in place of SGD. 

A drawback to \maml is that computing the ``meta-gradient" $\nabla_\phi \Lagr(f_{\phi'}, Q)$ requires calculating a second derivative, since the gradient must backpropagate through the sequence of updates made by $\mbox{Finetune}(\phi, d \:| S)$. 
Fortunately, in the same work where \cite{finn2017model} introduce \maml, they propose a first order approximation of \maml, \fomaml, which ignores these second derivative terms and performs nearly as well as the original method. We utilize \fomaml in our experiments to avoid memory issues associated with \maml.

%% file: sections/benchmark.tex
\begin{table*}[t!]
\begin{center}
\begin{tabular}{| l | c  c  c  c | c  c  c  c | c  c  c  c |}
\hline
\multirow{2}{*}{Split} & \multicolumn{4}{|c|}{\atis} & \multicolumn{4}{|c|}{\snips} & \multicolumn{4}{|c|}{\fbtop} \\
 & \#Utt & \#IC & \#SL & \#SV & \#Utt & \#IC & \#SL & \#SV & \#Utt & \#IC & \#SL & \#SV \\
\hline
Train & 4,373 & 5 & 116 & 461 & 8,230 & 4 & 33 & 8,549 & 20,345 & 7 & 38 & 5,574  \\
Dev   & 662   & 7 & 122 & 260 & - & - & - & -  & 4,333  & 5 & 33 & 2,228  \\
Test  & 829   & 7 & 128 & 258 & 6,254 & 3 & 20 & 7,567 & 4,426  & 6 & 39 & 1,341  \\
\hline
Total & 5,864 & 19 & 366 & 583 & 14,484 & 7 & 53 & 13,599 & 29,104 & 18 & 110 & 6821 \\
\hline
\end{tabular}
\caption{Statistics on utterance (Utt), intent (IC), slot label (SL), and slot value (SV) counts for \atis, \fbtop, and \snips few-shot train, development, and test splits as well as the full dataset, provided under the heading total.}
\label{tab:splits}
\end{center}
\end{table*}

\section{Few-shot IC/\slt Benchmark}
As there is no existing benchmark for \emph{few-shot IC/\slt}, we propose few-shot splits for the
Air Travel Information System 
(\atis, \citet{hemphill1990atis}), \snips \cite{coucke2018snips}, and Task Oriented Parsing (\fbtop, \cite{gupta-etal-2018-semantic-parsing}) datasets.
A \emph{few-shot IC/\slt} benchmark is beneficial for two reasons. 
Firstly, the benchmark evaluates generalization across multiple domains. 
Secondly, researchers can combine these datasets in the future to experiment with larger
settings of $n$-way during training and evaluation.%

\subsection{Datasets}
\atis is a well-known dataset for dialog system research, which comprises conversations from the \emph{airline} domain. 
\snips, on the other hand, is a public benchmark dataset developed by the Snips corporation
to evaluate the quality of IC and \slt services. The \snips dataset comprises multiple domains including music, media, and weather.
\fbtop, which pertains to navigation and event search, is unique in that 35\% of the utterances contain multiple, nested intent labels. 
These hierarchical intents require the use of specialized models. Therefore, we utilize only the remaining, non-hierarchical 65\% of utterances in \fbtop. 
To put the size and diversity of these datasets in context, we provide
utterance, intent, slot-label, and slot value counts for each dataset in table \ref{tab:splits}.

\subsection{Few-shot Splits}
We target train, development, and test split sizes of 70\%, 15\%, and 15\%, respectively.
However, the ICs in these datasets are highly imbalanced, which prevents us from hitting these targets exactly. Thereby, we manually select the ICs to include in each split.
For the \snips dataset, we choose not to form a development split because there are only 7 ICs in the \snips dataset, and
we require a minimum of 3 ICs per split. %
During preprocessing we modify slot label names by adding the associated IC as a prefix to each slot. This preprocessing step ensures that the slot labels are no longer pure named entities, but specific semantic roles in the context of particular intents. 
In table \ref{tab:splits}, we provide statistics on the few-shot splits for each dataset. 

%% file: sections/experiments.tex
\section{Experiments}

\subsection{Episode Construction}

For train and test episodes, we sample both the the number of classes in each episode, the ``way" $n$, 
and the number of examples to include for each sampled class $l$, the class ``shot" $k_l$, using the procedure put forward in \cite{triantafillou2019meta}.
By sampling the shot and way, we allow for unbalanced support sets and a variable number of classes per episode. 
These allowances are compatible with the large degree of class imbalances present in our benchmark, which would make it difficult to apply 
a fixed shot and way for all intents.

To construct an episode given a few-shot class split $L_{split}$, we first  sample the way $n$ uniformly from the range $[3, |L_{split}|]$. 
We then sample $n$ intent classes uniformly at random from $L_{split}$ to form $L$.
Next, we sample the query shot $k_q$ for the episodes as follows:
$$k_q = \min(10, (\min_{l \in L} \: [0.5 \: * |X_l|] ) )$$
where $X_l$ is the set of examples with class label $l$.
Given the query shot $k_q$, we compute the target support set size for the episode as:

$$ |S| \: = \min(K_{\max}, \sum_{l \in L} \lceil \beta \min(20, |X_l| - \: q) \rceil) $$

where $\beta$ is sampled uniformly from the range $(0, 1]$
and $K_{\max}$ is the maximum episode size.
Lastly, we sample the support shot $k_l$ for each class as:

$$ k_l = \min(\lfloor R_l * (|S| - |L|) \rfloor + 1,  |X_l| - \: q)$$

where $R_l$ is a noisy estimate of the normalized proportion of the dataset made up by class $l$, which we compute as follows:

$$R_l = \frac{\exp(\alpha_l) * |X_l| }{\sum_{l' \in L} \exp(\alpha_{l'}) * |X_{l'}|)}$$

The noise in our estimate of the proportion $R_l$ is introduced by sampling the value of $\alpha_l$ uniformly from the interval $[\log(0.5), \log(2))$.

\subsection{Episode Sizes}
We present IC/\slt results for two settings of maximum episode size, $K_{\max} = 20$ and $K_{\max} = 100$, in tables \ref{tab:icsmall}/\ref{tab:slsmall} and \ref{tab:icbig}/\ref{tab:slbig}, respectively. 
When the maximum episode size $K_{\max} = 20$, the average support set shot $k_l$ is 3.58 for \atis, 3.78 for \fbtop, and 5.22 for \snips.
In contrast, setting the maximum episode size to $K_{\max} = 100$ increases the average support set shot $k_l$ to 9.15 for \atis, 9.81 for \fbtop, and 10.83 for \snips. 

\subsection{Training Settings}
In our experiments, we consider two training settings. One in which we train on episodes, or batches in the case of our baseline, from a single dataset. And another, \emph{joint} training approach that randomly selects the dataset from which to sample a given episode/batch. After sampling an episode, we remove its contents from a buffer of available examples. If there are no longer enough examples in the buffer to create an episode, we refresh the buffer to contain all examples. 

\subsection{Network Architecture}
The network architectures we explore, depicted in Figure \ref{fig:architecture}, consist of an embedding layer,
a sequence encoder, and two output layers for slots and intents, respectively. 
Each architecture uses a different pre-trained embedding layer type, which are either non-contextual or contextual. We experiment with one non-contextual embedding, \glove word vectors \cite{pennington2014glove}, as well as two contextual embeddings, \glove concatenated with \elmo embeddings \cite{Peters:2018}, and \bert embeddings \cite{devlin2018bert}.
The sequence encoder is a bi-directional \lstm \cite{hochreiter1997long} with a 512-dimensional hidden state.
Output layers are fully connected and take concatenated forward and backward \lstm hidden states as input.
Pre-trained embeddings are kept frozen for training and adaptation. Attempts to fine-tune BERT led to inferior results.  
We refer to each architecture by its embedding type, namely \glove, \elmo, or \bert.

\subsection{Baseline}
We compare the performance of our approach against a \ft baseline, which
implements the domain adaptation
framework commonly applied to low resource IC/\slt \cite{goyal2018fast}. 
We pre-train the \ft baseline, either jointly or individually, on the classes in our training split(s).
Then at evaluation time, we freeze the pre-trained encoder 
and ``fine-tune" new output layers for the slots and intents included in the support set.
This fine-tuned model is then used to predict the intent and slots for each held out example in the query set. 

\subsection{Hyper-parameters}
We train all models using the Adam optimizer \cite{kingma2014adam}. 
We use the default learning rate of 0.001 for the baseline and prototypical networks.
For \fomaml we set the outer learning rate to 0.0029 and finetune for $d = 8$ steps with an inner learning rate of 0.01.
We pre-train the \ft baseline with a batch size of 512. %
At test time, we fine-tune the baseline for 10 steps on the support set.
We train the models without contextual embeddings (GloVe alone) for 50 epochs and those with contextual ELMo or BERT embeddings for 30 epochs because they exhibit faster convergence.

\subsection{Evaluation Metrics}
To assess the performance of our models,
we report the average IC accuracy
and slot F1 score over 100 episodes sampled from the test split of an individual dataset. 
We use the AllenNLP \cite{Gardner2017AllenNLP} CategoricalAccuracy implementation to compute IC Accuracy. 
And to compute slot F1 score, we use the seqeval library's span based F1 score implementation.\footnote{https://github.com/chakki-works/seqeval} 
The span based F1 score is a relatively harsh metric in the sense that a slot label prediction is only considered correct 
if the slot label and span exactly match the ground truth annotation.

%% file: sections/tables.tex
\begin{table*}[ht!]
\begin{center}
\small
\begin{tabular}{| l  l | c | c | c | c | c | c |}
\hline
\multirow{2}{*}{Embed.} & \multirow{2}{*}{Algorithm} & \multicolumn{6}{|c|}{IC Accuracy} \\
\cline{3-8}
& & \snips & \snips (joint) & \atis & \atis (joint) & \fbtop & \fbtop (joint) \\
\hline
GloVe & Fine-tune & 69.52 +/- 2.88          & 70.25 +/- 1.85          & 49.50 +/- 0.65          & 58.26 +/- 1.12          & 37.58 +/- 0.54          & 40.93 +/- 2.77 \\
GloVe & \fomaml   & 61.08 +/- 1.50          & 59.67 +/- 2.12          & 54.66 +/- 1.82          & 45.20 +/- 1.47          & 33.75 +/- 1.30          & 31.48 +/- 0.50 \\
GloVe & Proto     & 68.19 +/- 1.76          & 68.77 +/- 1.60          & \textbf{65.46 +/- 0.81} & \textbf{63.91 +/- 1.27} & 43.20 +/- 0.85          & 38.65 +/- 1.35 \\

\hline         
ELMo & Fine-tune & \textbf{85.53 +/- 0.35} & \textbf{87.64 +/- 0.73} & 49.25 +/- 0.74          & 58.69 +/- 1.56          & 45.49 +/- 0.61          & 47.63 +/- 2.75 \\
ELMo & \fomaml   & 78.90 +/- 0.77          & 78.86 +/- 1.31          & 53.90 +/- 0.96          & 52.47 +/- 2.86          & 38.67 +/- 1.02          & 36.49 +/- 0.99 \\
ELMo & Proto     & 83.54 +/- 0.40          & 85.75 +/- 1.57          & \textbf{65.95 +/- 2.29} & \textbf{65.19 +/- 1.29} & \textbf{50.57 +/- 2.81} & \textbf{50.64 +/- 2.72} \\
\hline      
BERT & Fine-tune & 76.04 +/- 8.84          & 77.53 +/- 5.69          & 43.76 +/- 4.61          & 50.73 +/- 3.86          & 39.21 +/- 3.09          & 40.86 +/- 3.75 \\
BERT & \fomaml   & 67.36 +/- 1.03          & 68.37 +/- 0.48          & 50.27 +/- 0.69          & 48.80 +/- 2.82          & 38.50 +/- 0.43          & 36.20 +/- 1.21 \\
BERT & Proto     & 81.39 +/- 1.85          & 81.44 +/- 2.91          & 58.84 +/- 1.33          & 58.82 +/- 1.55          & \textbf{52.76 +/- 2.26} & \textbf{52.64 +/- 2.58} \\
\hline
\end{tabular}
\caption{$K_{\max} = 20$ average IC accuracy on 100 test episodes from the \atis, \snips, or \fbtop datasets in the form mean $\pm$ standard deviation, computed over 3 random seeds, comparing GloVe, ELMo, and BERT model variants for both individual and \emph{joint} training, where we train on all training sets and test on a specific test set.}
\label{tab:icsmall}
\end{center}
\end{table*}

\begin{table*}[ht!]
\begin{center}
\small
\begin{tabular}{| l  l | c | c | c | c | c | c |}
\hline
\multirow{2}{*}{Embed.} & \multirow{2}{*}{Algorithm} & \multicolumn{6}{|c|}{IC Accuracy} \\
\cline{3-8}
& & \snips & \snips (joint) & \atis & \atis (joint) & \fbtop & \fbtop (joint) \\
\hline
GloVe & Fine-tune & 72.24 +/- 2.58          & 73.00 +/- 1.84          & 49.91 +/- 1.90          & 56.07 +/- 2.94          & 39.66 +/- 1.34          & 41.10 +/- 0.65 \\
GloVe & \fomaml   & 66.75 +/- 1.28          & 67.34 +/- 2.62          & 54.92 +/- 0.87          & 58.46 +/- 1.91          & 33.62 +/- 1.53          & 35.68 +/- 0.62 \\
GloVe & Proto     & 70.45 +/- 0.49          & 72.66 +/- 1.96          & \textbf{70.25 +/- 0.39} & 69.58 +/- 0.41          & 48.84 +/- 1.59          & 46.85 +/- 0.86 \\
                                 
\hline                           
ELMo & Fine-tune & \textbf{87.69 +/- 1.05} & \textbf{88.90 +/- 0.18} & 49.42 +/- 0.79          & 56.99 +/- 2.12          & 47.44 +/- 1.61          & 48.87 +/- 0.54 \\
ELMo & \fomaml   & 80.80 +/- 0.47          & 81.62 +/- 1.07          & 59.10 +/- 2.52          & 56.16 +/- 1.34          & 41.80 +/- 1.49          & 36.24 +/- 0.79 \\
ELMo & Proto     & \textbf{86.76 +/- 1.62} & 87.74 +/- 1.08          & \textbf{70.10 +/- 1.26} & \textbf{71.89 +/- 1.45} & 58.60 +/- 1.91          & 56.87 +/- 0.39 \\
                                    
\hline                              
BERT & Fine-tune & 76.66 +/- 8.68          & 79.53 +/- 4.25          & 44.08 +/- 6.05          & 49.71 +/- 3.84          & 40.05 +/- 2.35          & 40.46 +/- 1.74 \\
BERT & \fomaml   & 70.43 +/- 1.56          & 72.79 +/- 1.11          & 51.36 +/- 3.74          & 50.25 +/- 0.88          & 36.15 +/- 2.17          & 35.24 +/- 0.35 \\
BERT & Proto     & 83.51 +/- 0.88          & 86.29 +/- 1.09          & 66.89 +/- 2.31          & 65.70 +/- 2.31          & \textbf{61.30 +/- 0.32} & \textbf{62.51 +/- 1.79} \\
\hline
\end{tabular}
\caption{$K_{\max} = 100$ average IC accuracy on 100 test episodes from the \atis, \snips, or \fbtop datasets in the form mean $\pm$ standard deviation, computed over 3 random seeds, comparing GloVe, ELMo, and BERT model variants for both individual and \emph{joint} training, where we train on all training sets and test on a specific test set.}
\label{tab:icbig}
\end{center}
\end{table*}

\begin{table*}[ht!]
\begin{center}
\small
\begin{tabular}{| l  l | c | c | c | c | c | c |}
\hline
\multirow{2}{*}{Embed.} & \multirow{2}{*}{Algorithm} & \multicolumn{6}{|c|}{Slot F1 Measure} \\
\cline{3-8}
& & \snips & \snips (joint) & \atis & \atis (joint) & \fbtop & \fbtop (joint) \\
\hline
GloVe & Fine-tune & 6.72 +/- 1.24           &  6.68 +/- 0.40          &  2.57 +/- 1.21          & 13.22 +/- 1.07          &  0.90 +/- 0.51          & 0.76 +/- 0.21 \\
GloVe & \fomaml   & 14.07 +/- 1.01          & 12.91 +/- 0.43          & 18.44 +/- 0.91          & 16.91 +/- 0.32          &  5.34 +/- 0.43          & 9.22 +/- 1.03 \\
GloVe & Proto     & 29.63 +/- 0.75          & 27.75 +/- 2.52          & 31.19 +/- 1.15          & 38.45 +/- 0.97          & 10.65 +/- 0.83          & 18.55 +/- 0.35 \\
                                                      
\hline                                                
ELMo & Fine-tune & 22.02 +/- 1.13          & 16.00 +/- 2.07          &  7.47 +/- 2.60          &  7.19 +/- 1.71          &  1.26 +/- 0.46          & 1.17 +/- 0.32 \\
ELMo & \fomaml   & 33.81 +/- 0.33          & 32.82 +/- 0.84          & 27.58 +/- 1.25          & 24.45 +/- 1.20          & \textbf{22.35 +/- 1.23} & 15.53 +/- 0.64 \\
ELMo & Proto     & \textbf{59.88 +/- 0.53} & \textbf{59.73 +/- 1.72} & 33.97 +/- 0.38          & \textbf{40.90 +/- 2.21} & 20.12 +/- 0.25          & \textbf{28.97 +/- 0.82} \\
                           
\hline                     
BERT & Fine-tune & 12.47 +/- 0.31          &  8.75 +/- 0.28          &  9.24 +/- 1.67          & 15.93 +/- 3.10          &  3.15 +/- 0.28          & 1.08 +/- 0.30 \\
BERT & \fomaml   & 12.72 +/- 0.12          & 13.28 +/- 0.53          & 18.91 +/- 1.01          & 16.05 +/- 0.32          &  5.93 +/- 0.43          & 8.23 +/- 0.81 \\
BERT & Proto     & 42.09 +/- 1.11          & 43.77 +/- 0.54          & \textbf{37.61 +/- 0.82} & \textbf{39.27 +/- 1.84} & 20.81 +/- 0.40          & \textbf{28.24 +/- 0.53} \\
\hline
\end{tabular}
\caption{$K_{\max} = 20$ average Slot F1 score on 100 test episodes from the \atis, \snips, or \fbtop datasets in the form mean $\pm$ standard deviation, computed over 3 random seeds, comparing GloVe, ELMo, and BERT model variants for both individual and \emph{joint} training, where we train on all training sets and test on a specific test set.}
\label{tab:slsmall}
\end{center}
\end{table*}

\begin{table*}[ht!]
\begin{center}
\small
\begin{tabular}{| l  l | c | c | c | c | c | c |}
\hline
\multirow{2}{*}{Embed.} & \multirow{2}{*}{Algorithm} & \multicolumn{6}{|c|}{Slot F1 Measure} \\
\cline{3-8}
& & \snips & \snips (joint) & \atis & \atis (joint) & \fbtop & \fbtop (joint) \\
\hline
GloVe & Fine-tune &  7.06 +/- 1.87          &  7.76 +/- 0.91          &  2.72 +/- 1.65          & 17.20 +/- 3.03          &  1.26 +/- 0.44           &  0.67 +/- 0.33 \\
GloVe & \fomaml   & 16.77 +/- 0.67          & 16.53 +/- 0.32          & 17.80 +/- 0.42          & 23.33 +/- 2.89          &  4.11 +/- 0.81           &  9.89 +/- 1.13 \\
GloVe & Proto     & 31.57 +/- 1.28          & 31.17 +/- 1.31          & 31.32 +/- 2.79          & 41.07 +/- 1.14          & 9.99 +/- 1.08            & 18.93 +/- 0.77 \\
                                                      
\hline                                                
ELMo & Fine-tune & 22.37 +/- 0.91          & 17.09 +/- 2.57          &  8.93 +/- 2.86          & 11.09 +/- 2.00          &  2.04 +/- 0.41           &  1.03 +/- 0.24 \\
ELMo & \fomaml   & 36.10 +/- 1.49          & 37.33 +/- 0.24          & 26.91 +/- 2.64          & 26.37 +/- 0.15          & 18.32 +/- 0.52           & 16.55 +/- 0.79 \\
ELMo & Proto     & \textbf{62.71 +/- 0.40} & \textbf{62.14 +/- 0.75} & 35.20 +/- 2.46          & \textbf{41.28 +/- 2.73} & \textbf{18.44 +/- 2.41}  & \textbf{28.33 +/- 1.33} \\
                                    
\hline                              
BERT & Fine-tune & 14.71 +/- 0.43          & 10.50 +/- 0.90          & 11.53 +/- 1.46          & 20.41 +/- 1.85          &  4.98 +/- 0.66           &  1.48 +/- 0.85 \\
BERT & \fomaml   & 14.99 +/- 1.29          & 15.83 +/- 0.94          & 17.68 +/- 2.42          & 17.11 +/- 1.31          &  3.37 +/- 0.36           & 10.58 +/- 0.45 \\
BERT & Proto     & 46.50 +/- 0.75          & 48.77 +/- 0.71          & \textbf{40.63 +/- 3.37} & \textbf{43.10 +/- 1.76} & \textbf{20.58 +/- 2.27}  & \textbf{28.92 +/- 1.09} \\
\hline
\end{tabular}
\caption{$K_{\max} = 100$ average Slot F1 score on 100 test episodes from the \atis, \snips, or \fbtop datasets in the form mean $\pm$ standard deviation, computed over 3 random seeds, comparing GloVe, ELMo, and BERT model variants for both individual and \emph{joint} training, where we train on all training sets and test on a specific test set.}
\label{tab:slbig}
\end{center}
\end{table*}

%% file: sections/results.tex
\section{Results}\label{sec:results}

\subsection{Few-shot Learning Algorithms}

\paragraph{Prototypical networks} 
Considering both IC and \slt tasks, prototypical networks is 
the best performing algorithm. 
The most successful variant of prototypical networks, Proto \elmo + \emph{joint} training, 
obtains absolute improvements over the \ft \elmo + \emph{joint} training baseline
of up to 6\% IC accuracy and 43 slot F1 points for $K_{max} = 20$, and 14\% IC accuracy and 45 slot F1 points for $K_{max} = 100$. 
The one case in which Proto \elmo + \emph{joint} training does worse than the baseline is on \snips IC, but these losses are all under 2\%.   

\paragraph{\fomaml} 
The results for \fomaml are more mixed in terms of IC and \slt 
performance relative to the baseline. The best \fomaml variant, \fomaml \elmo, underperforms \ft \elmo on \snips and \fbtop IC by up to 6\%. 
Yet \fomaml improves IC accuracy by 4\% ($K_{max} = 20$) to 9\% ($K_{max} = 100$) on \atis. 
\fomaml \elmo consistently outperforms \ft \elmo on \slt for all datasets, generating gains of 11$\sim$21 F1 points for $K_{max} = 20$  and 13$\sim$17 F1 points for $K_{max} = 100$. Notably, \bert and \fomaml in combination do not work well. Specifically, the \slt performance of \fomaml \bert is comparable to, or worse than, \fomaml \glove on all datasets for both $K_{max} = 20$ and $K_{max} = 100$. 

\subsection{Model Variants}

\paragraph{Non-contextual Pretrained Embeddings} 
The \glove model architecture, which uses \glove alone, does not perform as well as \elmo or \bert. 
On average over experimental settings, the \glove variant of the winning algorithm has 10\% lower IC Accuracy and 16 point lower slot F1 score than the winning algorithm paired with the best model. 
Note that an experimental setting here refers to a combination of dataset, value of $K_{max}$, and use of individual or joint training.
Somewhat surprisingly, \glove performs nearly as well as \elmo and even better than \bert on \atis IC. 
We speculate that \atis IC does not benefit as much from the use of \elmo or \bert because \atis carrier phrases are less diverse, as evidenced by the smaller number of unique carrier phrases in the \atis test set (527) compared to \snips (3,718) and \fbtop (4,153).

\paragraph{Contextual Pretrained Embeddings} 
A priori, it is reasonable to suspect that the performance gain obtained by our few-shot learning algorithms could be dwarfed by the benefit of using a large, pre-trained model like \elmo or \bert. However, our experimental results suggest that the use of pre-trained language models is complementary to our approach, in most cases. 
For example, \elmo increases the slot F1 score of \fomaml from 14.07 to 33.81 and boosts the slot F1 of prototypical networks from 31.57 to 62.71 on the \snips dataset for $K_{max} = 100$. Similarly, when $K_{max} = 20$, \bert improves \fomaml and prototypical networks \fbtop IC accuracy from 33.75\% to 38.50\% and from 43.20\% to 52.76\%, respectively.
In aggregate, we find \elmo outperforms \bert. We quantify this via the average absolute improvement \elmo obtains over \bert when both models use the winning algorithm for a given dataset and training setting. %
On average, \elmo improves IC accuracy over \bert by 2\% for $K_{max} = 20$ and 1\% for $K_{max} = 100$. With respect to slot F1 score, \elmo produces an average gain over \bert of 5 F1 points for $K_{max} = 20$ and 3 F1 points for $K_{max} = 100$. 
This is consistent with previous findings in \cite{peters-etal-2019-tune} that \elmo can outperform \bert on certain tasks when the models are kept frozen and not fine-tuned. 

\subsection{Joint Training}
Few-shot learning algorithms are in essence learning to learn new classes. 
Therefore, these algorithms should be better suited to leverage  
a diverse training dataset to improve generalization. 
We test this hypothesis by jointly training each approach on all three datasets.
Our results demonstrate that joint training has little effect on IC Accuracy; 
however, it improves the \slt performance of prototypical networks, particularly on \atis and \fbtop.
Joint training increases Prototypical networks average slot F1 score, computed over datasets and model variants, by 4.41 points from 31.77 to 36.18 for $K_{max} = 20$ and by 5.20 points from 32.99 to 38.19 when $K_{max} = 100$. In comparison, Fine-tune obtains much smaller average absolute improvements, 0.55 F1 points and 1.29 F1 points for $K_{max} = 20$ and $K_{max} = 100$, respectively.

%% file: sections/conclusion.tex
\section{Conclusion}
This work shows the benefit of applying few-shot learning techniques to \emph{few-shot IC/\slt}.
Specifically, our extension of prototypical networks for joint IC and \slt consistently outperforms a fine-tuning based method with respect to both IC Accuracy and slot F1 score. 
The use of this prototypical approach in combination with pre-trained language models, such as ELMo, generates additional performance improvements, especially on the \slt task.
While our contribution is a step toward the creation of more sample efficient IC/\slt models, there is still substantial work to be done in pursuit of this goal, especially in the creation of larger \emph{few-shot IC/\slt} benchmarks. We encourage the creation of a large scale IC and \slt dataset to test how these methods scale with larger episode sizes and view this direction as a high leverage way to further this line of research.